\pdfoutput=1

\documentclass[11pt]{article}

\usepackage[]{acl}

\usepackage{times}
\usepackage{latexsym}

\usepackage[T1]{fontenc}

\usepackage[utf8]{inputenc}

\usepackage{microtype}

\usepackage{amsmath}
\usepackage{amssymb}
\usepackage{multirow}
\usepackage{graphicx}
\usepackage{subfigure}
\usepackage{threeparttable}
\usepackage{algpseudocode}
\usepackage{algorithm}
\usepackage{color}

\def\figlabel#1{\label{fig:#1}\label{p:#1}}

\def\tabref#1{Table~\ref{tab:#1}}

\def\tablabel#1{\label{tab:#1}\label{p:#1}}

\newcounter{notecounter}

\usepackage{amsfonts} 
\usepackage{xspace} 
\usepackage{textcomp}
\def\md{RMLNMT\xspace}
\title{Improving Both Domain Robustness and Domain Adaptability \\in Machine Translation}

\author{Wen Lai$^1$ \and Jindřich Libovický$^2$ \and Alexander Fraser$^1$ \\
        $^1$ Center for Information and Language Processing, LMU Munich, Germany \\ 
        $^2$ Faculty of Mathematics and Physics, Charles Univeristy, Prague, Czech Republic \\
        \tt \{lavine, fraser\}@cis.lmu.de libovicky@ufal.mff.cuni.cz}

\begin{document}
\maketitle
\begin{abstract}
We consider two problems of NMT domain adaptation using meta-learning.
First, we want to reach domain \emph{robustness}, i.e., 
we want to reach high quality on both domains seen in the training data and unseen domains.
Second, we want our systems to be \emph{adaptive}, i.e., making it possible to finetune systems with just hundreds of in-domain parallel sentences.
We study the domain adaptability of meta-learning when improving the domain robustness of the model.
In this paper, we propose a novel approach, \textbf{RMLNMT} (\textbf{R}obust \textbf{M}eta-\textbf{L}earning Framework for \textbf{N}eural \textbf{M}achine \textbf{T}ranslation Domain Adaptation), which improves the robustness of existing meta-learning models.
More specifically, we show how to use a domain classifier in curriculum learning and we integrate the word-level domain mixing model into the meta-learning framework with a balanced sampling strategy.
Experiments on English$\rightarrow$German and English$\rightarrow$Chinese translation show that \md improves in terms of both domain robustness and domain adaptability in seen and unseen domains\footnote{Our source code is available at \href{https://github.com/lavine-lmu/RMLNMT}{https://github.com/lavine-lmu/RMLNMT}}.
\end{abstract}

\section{Introduction}
The success of Neural Machine Translation (NMT; \citealp{bahdanau2014neural,vaswani2017attention}) heavily relies on large-scale high-quality parallel data, which is difficult to obtain in some domains.
We study two major problems in NMT domain adaptation.
First, models should work well on both seen domains (the domains in the training data) and unseen domains (domains which do not occur in the training data). We call this property \textit{domain robustness}.
Second, with just hundreds of in-domain sentences, we want to be able to quickly adapt to a new domain. We call this property \textit{domain adaptability}.
Previous work on NMT domain adaptation has usually focused on only one aspect of domain adaptation at the expense of the other one, and our motivation is to consider both of the two properties.

There are a few works attempting to solve domain adaptability.
The most basic approach is \textit{fine-tuning}, in which an out-of-domain model is continually trained on in-domain data \citep{freitag2016fast,dakwale2017finetuning}.
Although fine-tuning is effective, it can suffer from so-called catastrophic forgetting \cite{CFFRENCH1999}, resulting in deteriorated model performance in general domains \cite{thompson-etal-2019-overcoming}.
Another efficient method is \textit{Meta-Learning} \cite{hospedales2021meta}, which trains models which can be later rapidly adapted to new scenarios using only a small amount of data.
It works for many natural language processing (NLP) tasks \citep{gu-etal-2018-meta,qian-yu-2019-domain,yu-etal-2020-hypernymy,bansal-etal-2020-self, wang-etal-2021-variance,du-etal-2021-meta}, especially in low-resource scenarios \citep{dou-etal-2019-investigating, yin2020meta}.
As a result, meta-learning is often used for NMT domain adaptation. For example, \citet{sharaf-etal-2020-meta} and \citet{li2020metamt} fast adapt NMT models to new domains with meta-learning using a small amount of training data.
\citet{Zhan_Liu_Wong_Chao_2021} improve meta-learning-based NMT models with a curriculum-based \cite{bengio2009curriculum} sampling strategy.
Meta-learning works well for adapting to new domains, however, previous work tends to neglect the problem of robustness towards domains unseen at training time.

\citet{muller2020domain} defined the concept of domain robustness and propose to improve the domain robustness by subword regularization \cite{kudo-2018-subword}, defensive distillation \cite{papernot2016distillation}, reconstruction \cite{tu2017neural} and neural noisy channel reranking \cite{yee-etal-2019-simple}.
\citet{jiang-etal-2020-multi-domain} proposed using individual modules for each domain with a word-level domain mixing strategy, which they showed has domain robustness on seen domains.
The work on domain robustness, however, tends to neglect the adaptability of the models for new domains.

To address both domain adaptability and domain robustness at the same time, we propose \md (robust meta-learning NMT), a more robust meta-learning-based NMT domain adaptation framework.
We first train a word-level domain mixing model to improve the robustness on seen domains, and show that, surprisingly, this improves robustness on unseen domains as well.
Then, we train a domain classifier based on BERT \cite{devlin-etal-2019-bert} to score training sentences; the score measures similarity between out-of-domain and general-domain sentences.
This score is used to determine a curriculum to improve the meta-learning process.
Finally, we improve domain adaptability by integrating the domain-mixing model into a meta-learning framework with the domain classifier using a balanced sampling strategy.

In summary, we make the following contributions:
i) we propose \md, which shows better domain robustness and domain adaptability than all previous baseline systems;
ii) we show that unseen domains can be very effectively handled with domain-robust models, even though post-hoc adaptation with domain-specific data still delivers the best overall translation quality;
iii) Experiments on English$\rightarrow$German and English$\rightarrow$Chinese translation tasks show the effectiveness of \md.
To the best of our knowledge, this is the first work that considers both domain adaptability and domain robustness in NMT domain adaptation, a combination which we suggest the community pay more attention to.

\section{Preliminaries}

\paragraph{Neural Machine Translation.}
The goal of the NMT model is to model the conditional distribution of translated sentence $y=(y_{1},...,y_{n})$ given a source sentence $x=(x_{1},...,x_{m})$.
Current state-of-art NMT models (Transformers; \citealp{vaswani2017attention}) model the multi-head attention mechanism to focus on information in different representation subspaces from different positions
$$ \operatorname{MultiHead}(Q,K,V) = \operatorname{Concat}\left({h}_{1}, \ldots, {h}_{\mathrm{h}}\right)W^{O} $$
$$h_{i}=\operatorname{Attention}\left(Q W_{i}^{Q}, K W_{i}^{K}, V W_{i}^{V}\right),$$
where $W_{i}^{Q}, W_{i}^{K}, W_{i}^{V} \in \mathbb{R}^{d \times d / m}$ and $W^{O} \in \mathbb{R}^{d \times d}$.
For the $i$-th head $h_{i}$, $m$ is the number of heads, and $d$ is the dimension of the model output.
In some of our experiments (see Section~\ref{sec:word-domain-mixing}), we modify the multihead attention to do domain mixing \cite{jiang-etal-2020-multi-domain}.

\paragraph{Meta-learning for NMT.}
The goal of Meta-Learning is training a teacher model that using previous experience can be better finetuned for new tasks, including handling different domains  in NMT domain adaptation \citep{gu-etal-2018-meta, sharaf-etal-2020-meta, Zhan_Liu_Wong_Chao_2021}.
The idea of NMT domain adaptation with meta-learning is to use a small set of source tasks $\left\{\mathcal{T}_{1}, \ldots, \mathcal{T}_{n}\right\}$
(which correspond to domains) to find the initialization of model parameters $\theta$ from which finetuning for task $\mathcal{T}_{0}$ would require only a small number of training examples.
These meta-learning algorithms consist of three main steps:
(i) split the seen domain corpus into small tasks $\mathcal{T}$ containing a small amount of data as $\mathcal{D}_{\text {meta-train}}$ and $\mathcal{D}_{\text {meta-test}}$ to simulate the low-resource scenarios. Data for each task $\mathcal{T}_{i}$ is decomposed into two sub-sets: a support set $\mathcal{T}_{support}$ used for training the model and a query set $\mathcal{T}_{query}$ used for evaluating the model;
(ii) leverage a meta-learning policy to adapt model parameters to different small tasks using $\mathcal{D}_{\text {meta-train}}$ datasets. We use MAML, proposed by \citet{finn2017model}, to create adaptable NMT systems which will be useful for different domains;
(iii) finetune the model using the support set of $\mathcal{D}_{\text {meta-test}}$.

\begin{figure*}
	\includegraphics[width=1\linewidth]{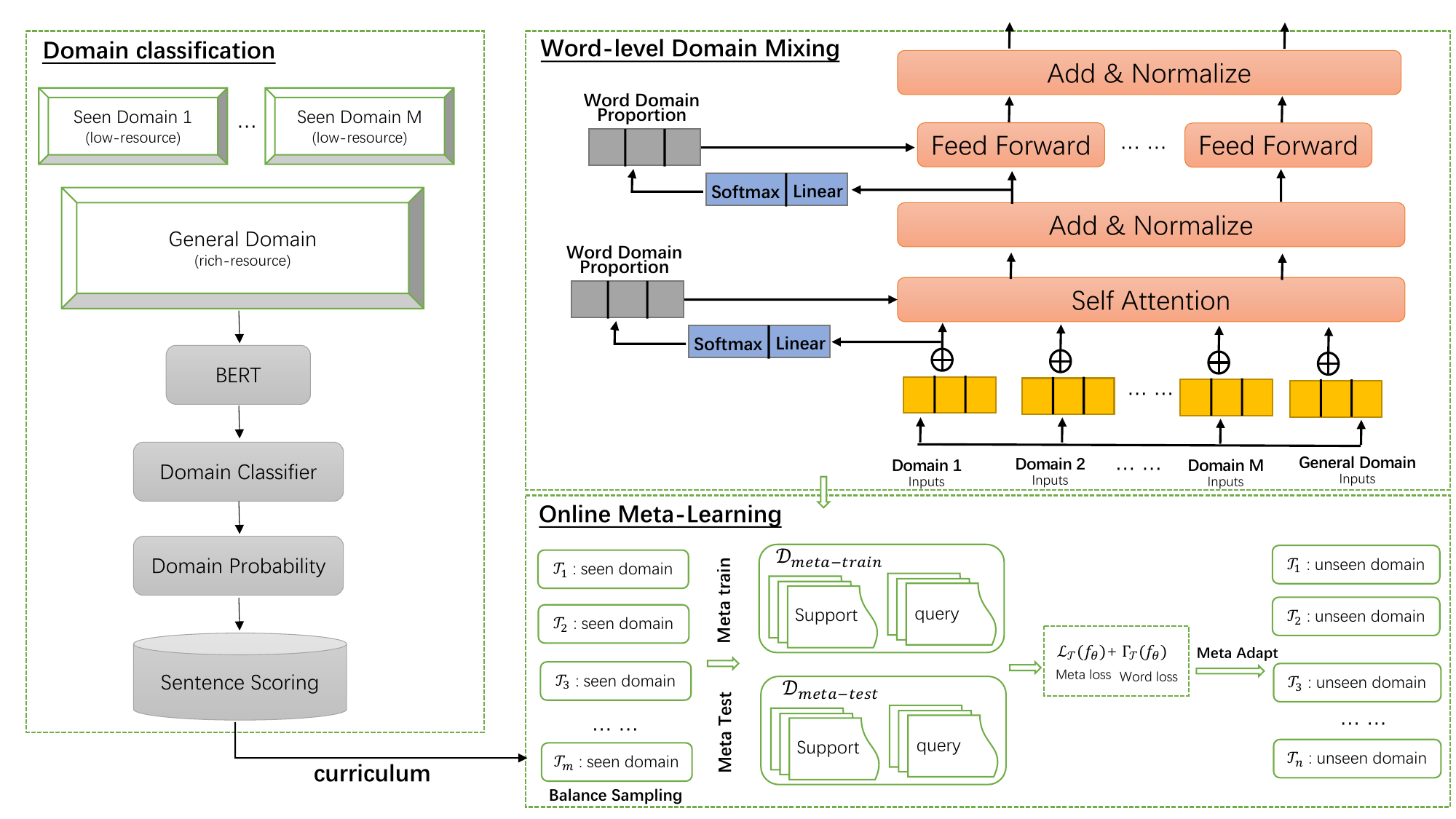}
	\caption{\label{fig:models}Method overview.
	The whole procedure mainly consists of three parts: domain classification, word-level domain mixing and online meta-learning.}
\end{figure*}

\section{Method}
In our initial experiments, we observed that the standard meta-learning approach for NMT domain adaptation sacrifices the domain robustness on seen domains in order to improve the domain adaptability on unseen domains.
To address these issues, we propose a novel approach, \md, which combines meta-learning with a word-level domain-mixing system (for improving domain robustness) in a single model.
\md consists of three parts: Word-Level Domain Mixing, Domain Classification, and Online Meta-Learning. Figure~\ref{fig:models} illustrates \md.

\subsection{Word-level Domain Mixing}\label{sec:word-domain-mixing}

In order to improve the robustness of NMT domain adaptation, we follow the approach of \citet{jiang-etal-2020-multi-domain} and train a word-level layer-wise domain mixing NMT model.

\paragraph{Domain Proportion.}
From a sentence-level perspective (i.e., the classifier-based curriculum step), each sentence has a domain label. However, the domain of a word in the sentence is not necessarily consistent with the sentence domain. E.g., the word \textit{doctor} can have a different meaning in the medical domain and the academic domain.
More specifically, for $k$ domains, the embedding $\mathbf{w}\in\mathbb{R}^{d}$ of a word, and a matrix $R \in \mathbb{R}^{k \times d}$, the domain proportion of the word is represented by a smoothed softmax function as:
$$\Phi(\mathbf{w})=(1-\epsilon) \cdot \operatorname{softmax}(R \mathbf{w})+\epsilon / k,$$
\noindent where $\epsilon \in(0,1)$ is a smoothing parameter to 
prevent the output
of $\Phi(\mathbf{w})$ from collapsing towards $0$ or $1$.

\paragraph{Domain Mixing.}
Following \citet{jiang-etal-2020-multi-domain}, each domain has its own multi-head attention modules.
Therefore, we can integrate the domain proportion of each word into its multi-head attention module.
Specifically, we take the weighted average of the linear transformation based on the domain proportion $\Phi$.
For example, we consider the point-wise linear transformation $\left\{W_{i, V, j}\right\}_{j=1}^{k}$ on the $t$-th word of the input, $V_{t}$, of all domains. 
The mixed linear transformation can be written as
$$
\bar{V}_{i, t}=\sum_{j=1}^{k} V_{t}^{\top} W_{i, V, j} \Phi_{V, j}\left(V_{t}\right),
$$
where $\Phi_{V, j}\left(V_{t}\right)$ denotes the $j$-th entry of $\Phi_{V}\left(V_{t}\right)$, and $\Phi_{V}$ is the domain proportion layer related to $V$. 
For other linear transformations, we apply the domain mixing scheme in the same way for all attention layers and the fully-connected layers.

\paragraph{Training.}
The model can be efficiently trained by minimizing a composite loss:
$$L^{*}=L_{\mathrm{gen}}(\theta)+L_{\mathrm{mix}}(\theta),$$
where $\theta$ contains the parameter in encoder, decoder and domain proportion.
$L_{\text {gen }}(\theta)$ denotes the cross-entropy loss over training data $\left\{\mathbf{x}_{i}, \mathbf{y}_{i}\right\}_{i=1}^{n}$ and $L_{\mathrm{mix}}(\theta)$ denotes the cross-entropy loss over the words/domain labels. For $L_{\mathrm{mix}}(\theta)$, we compute the cross-entropy loss of its domain proportion $\Phi(\mathbf{w})$ as $-\log \left(\Phi_{J}(\mathbf{w})\right)$, which take $J$ as the domain label. Hence, $L_{\mathrm{mix}}(\theta)$ is computed as the sum of the cross-entropy loss over all such pairs of word labels of the training data.

\subsection{Domain Classification}\label{sec:cls_domain}
Domain similarity has been successfully applied in NMT domain adaptation.
\citet{moore-lewis-2010-intelligent} calculate cross-entropy scores with a language model to represent the domain similarity.
\citet{riess21:mtsummit} leverage simple classifiers to compute similarity scores; these scores are more effective than scores from language models for NMT domain adaptation.
Motivated by \citet{riess21:mtsummit}, we compute domain similarity using a sentence-level classifier, but in contrast with previous work, we based our classifier on a pre-trained language model.
Given $k$ domain corpora (one general domain corpus and $n$ out-of-domain corpora), we trained a sentence classification model $M$ based on BERT \cite{devlin-etal-2019-bert}.
For a sentence $x$ with a domain label $L_x$, a simple softmax is added to the top of the model $M$ to predict the domain probability of sentence $x$:
$$
P(x\mid h)=\operatorname{softmax}(W h),
$$
where $W$ is the parameter matrix of $M$ and $h$ is the hidden state of $M$.
$P(x \mid h)$ is a probability set, which contains $k$ probability scores indicating the similarity of sentence $x$ to each domain.
We finally select the probability of the general domain (from $k$ probability scores) as the score of the sentence $x$ and use this score as the curriculum to split the task in meta-learning (see more details in Section~\ref{sec:meta-learning}).
A higher score indicates that the sentence is more similar to the general domain, so we will select it earlier.

\subsection{Online Meta-Learning}\label{sec:meta-learning}
After training the word-level domain mixing NMT model, we use it as a teacher model to initialize the meta-learning process.
Algorithm~\ref{alg} shows the complete algorithm.

\paragraph{Split Tasks.} 
\citet{Zhan_Liu_Wong_Chao_2021} propose a curriculum-based task splitting strategy, which uses divergence scores computed by a language model as the curriculum to split the corpus into small tasks.
We follow a similar idea, but propose to use predictions from a domain classifier as the criterion for splitting the data.
Concretely, we first train a domain classifier with BERT; the classifier scores sentences, indicating domain similarity between an in-domain sentence and a general domain sentence (see Section~\ref{sec:cls_domain}).
The tasks are then split according to the scores; sentences more similar to the general domain sentences are selected in early tasks.

\begin{figure}[t]
    \includegraphics[width=1\linewidth,trim=0pt 5pt 0pt 25pt]{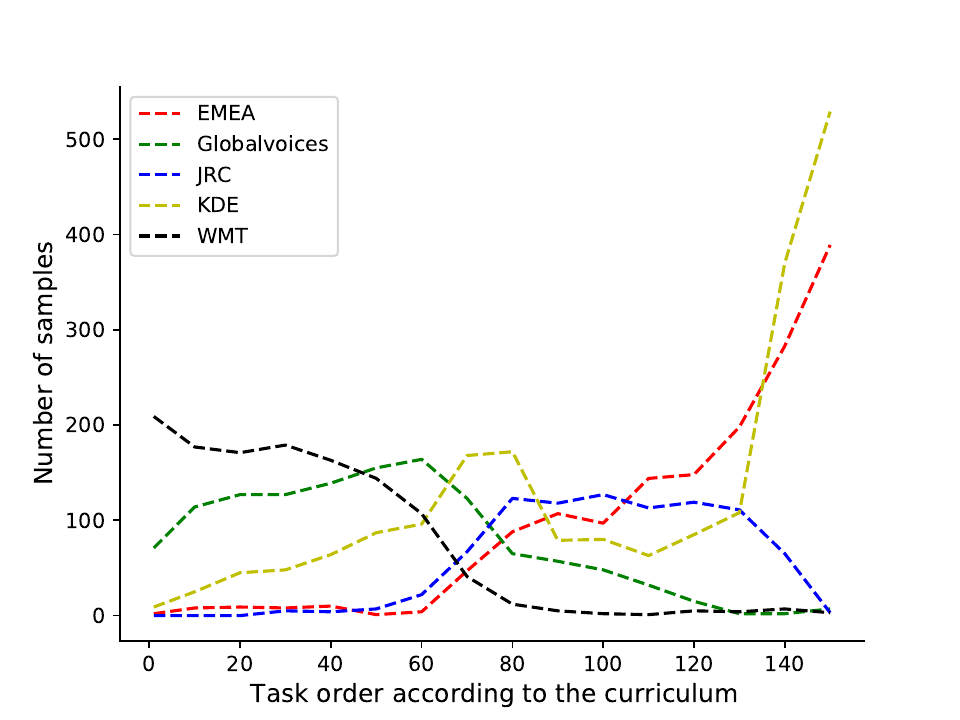}
	\caption{\label{fig:sample_sts} The statistic of samples in the task for the tokenization-based 
 splitting strategy. More general domains are on the left and the more distinctive domains are on the right.}
\end{figure}

\paragraph{Balanced Sampling.}
Previous meta-learning approaches \citep{sharaf-etal-2020-meta, Zhan_Liu_Wong_Chao_2021} are based on token-size based sampling, which uses $8k$ or $16k$ token sizes split into many small tasks.
However, the splitting process for the domain is not balanced, since some tasks did not contain all seen domains, especially in the early tasks.
As we can see in Figure~\ref{fig:sample_sts}, the token-based splitting methods usually allocate more samples on domain-similar domains (\textit{WMT}, \textit{Globalvoices}) and allocate small samples on domain-distant domains (\textit{EMEA}, \textit{JRC}) in the sampling of early tasks.
This can cause problems in our method since the model architecture is dynamically changing according to the number of domains (see more details in Section~\ref{sec:word-domain-mixing}).

To address these issues, we sample the data uniformly from the domains to compensate for imbalanced domain distributions based on domain classifier scores.

\paragraph{Meta-Training.}
Following the balanced sampling, the process of meta-training is to update the current model parameter on $\mathcal{T}_{support}$ from $\theta$ to $\theta^{\prime}$, and then evaluate on $\mathcal{T}_{query}$.
The model parameter $\theta^{\prime}$ is updated to minimize the meta-learning loss through MAML.

Given a pre-trained model $f_{\theta}$ (initialized with parameters $\theta$ trained on word-level domain mixing) and the meta-train data $\mathcal{D}_{\text {meta-train}}$, for each task $\mathcal{T}$, we learn to use one gradient update to update the model parameters from $\theta$ to $\theta^{\prime}$ as follows:
$$
    \theta^{\prime}=\theta-\alpha \nabla_{\theta} L_{\mathcal{T}}\left(f_{\theta}\right)
$$
where $\alpha$ is the learning rate and $L$ is the loss function.
In our methods, we consider both the traditional sentence-level meta-learning loss $\mathcal{L}_{\mathcal{T}}\left(f_{\theta}\right)$ and the word-level loss $\Gamma_{\mathcal{T}}\left(f_{\theta}\right)$ ($L^{*}$ of $\mathcal{T}$) calculated from the word-level domain mixing pre-trained model.
More formally, the loss is updated as follows:
$$
L_{\mathcal{T}}\left(f_{\theta}\right) = \mathcal{L}_{\mathcal{T}}\left(f_{\theta}\right) +  \Gamma_{\mathcal{T}}\left(f_{\theta}\right).
$$

Note that the meta-training phase is not adapted to a specific domain, so it can be used as a metric to evaluate the domain robustness of the model.

\paragraph{Meta-Adaptation.} After the meta-training phase, the parameters are updated to adapt to each domain using the small \emph{support set} of $\mathcal{D}_{\text {meta-test}}$ corpus to simulate the low-resource scenarios.
Then performance is evaluated on the \emph{query set} of $\mathcal{D}_{\text {meta-test}}$.

\begin{algorithm}[t]
\caption{RMLNMT (Robust Meta-Learning NMT Domain Adaptation)}\label{alg}
\begin{algorithmic}[1]
\Require Domain classifier model $cls$; Pretrained domain-mixing model $\theta$;
\State Score the sentence in $\mathcal{D}_{\text {meta-train}}$ using $cls$
\For {$N$ epochs}
    \State Split corpus into $n$ tasks based on step 1
    \State Balance sample through all tasks
    \For{task $\mathcal{T}_i$, $i=1\ldots n$}
        \State Evaluate loss $L_{\mathcal{T}}\left(f_{\theta}\right) $
        
        \quad\quad $= \mathcal{L}_{\mathcal{T}_{i}}\left(f_{\theta}\right) +  \Gamma_{\mathcal{T}_{i}}\left(f_{\theta}\right)$ on support set
        \State Update the gradient with parameters
        
        \quad\quad $\theta^{\prime}=\theta-\alpha \nabla_{\theta} L_{\mathcal{T}}\left(f_{\theta}\right)$
    \EndFor
    \State Update the gradient with parameters
    
    \quad $\theta=\theta-\beta \nabla_{\theta} L_{\mathcal{T}}\left(f_{\theta^{\prime}}\right)$ on query set
\EndFor
\State \textbf{return} RMLNMT model parameter $\theta$
\end{algorithmic}
\end{algorithm}

\begin{table*}[t]
\centering
\resizebox{\textwidth}{!}{
\begin{tabular}{ll|ccccc|ccccc}
\hline
& \multirow{2}{*}{\textbf{Models}} & \multicolumn{5}{c}{\textbf{Unseen}} &\multicolumn{5}{c}{\textbf{Seen}} \\
\cline{3-12}
& & Covid & Bible & Books & ECB & TED & EMEA & Globalvoices & JRC & KDE & WMT \\
\hline
1 & \textbf{Vanilla} & 24.34 & \textbf{12.08} & 12.61 & 29.96 & 27.89 & 37.27 & 24.19 & 39.84 & 27.75 & 27.38 \\
\hline
2 & \textbf{Vanilla + tag} & 24.86 & 12.04 & 12.46 & 30.03 & 27.93 & 38.37 & 24.56 & 40.75 & 28.23 & 27.26 \\
\hline
3 & \textbf{Meta-MT w/o FT} & 23.69 & 11.07 & 12.10 & 29.04 & 26.86 & 30.94 & 23.73 & 38.82 & 23.04 & 26.13 \\
\hline
4 & \textbf{Meta-Curriculum (LM) w/o FT} & 23.70 & 11.16 & 12.24 & 28.22 & 27.21 & 33.49 & 24.27 & 39.21 & 27.60 & 25.83 \\
\hline
5 & \textbf{RMLNMT w/o FT} & \textbf{25.48} & 11.48 & \textbf{13.11} & \textbf{31.42} & \textbf{28.05} & \textbf{47.00} & \textbf{26.35} & \textbf{51.13} & \textbf{32.80} & \textbf{28.37} \\
\hline
\end{tabular}
}
\caption{\label{tab:en2de_res_domain_robustness}Domain Robustness: BLEU scores on the English $\rightarrow$ German translation task. \textit{w/o} denotes the meta-learning systems without fine-tuning, FT denotes fine-tuning. Best results are highlighted in bold.
}
\figlabel{tab:en2de_res_domain_robustness}
\end{table*}

\begin{table*}[t]
\centering
\resizebox{\textwidth}{!}{
\begin{tabular}{ll|ccccc|ccccc}
\hline
& \multirow{2}{*}{\textbf{Models}} & \multicolumn{5}{c}{\textbf{Unseen}} &\multicolumn{5}{c}{\textbf{Seen}} \\
\cline{3-12}
& & Covid & Bible & Books & ECB & TED & EMEA & Globalvoices & JRC & KDE & WMT \\
\hline
1 & \textbf{Plain FT} & 24.81 & 12.61 & 12.78 & 30.48 & 28.36 & 37.26 & 24.26 & 40.02 & 27.99 & 27.31 \\
\hline
2 & \textbf{Plain FT + tag} & 25.31 & 12.57 & 12.83 & 30.57 & 28.39 & 39.54 & 24.91 & 41.51 & 29.14 & 27.58 \\
\hline
3 & \textbf{Meta-MT + FT} & 25.83 & 14.20 & 13.39 & 30.36 & 28.57 & 34.69 & 24.64 & 39.15 & 27.47 & 26.38 \\
\hline
4 & \textbf{Meta-Curriculum (LM) + FT} & \textbf{26.66} & 14.37 & 13.70 & 30.41 & 28.97 & 34.00 & 24.72 & 39.61 & 27.37 & 26.68 \\
\hline
5 & \textbf{RMLNMT + FT} & 26.53 & \textbf{15.37} & \textbf{13.72} & \textbf{31.97} & \textbf{29.47} & \textbf{47.02} & \textbf{26.55} & \textbf{51.13} & \textbf{32.88} & \textbf{28.37} \\
\hline
\end{tabular}
}
\caption{\label{tab:en2de_res_domain_adapt}Domain Adaptability: BLEU scores on the English $\rightarrow$ German translation task.
}
\figlabel{tab:en2de_res_domain_adapt}
\end{table*}

\begin{table*}
\centering
\resizebox{\textwidth}{!}{
\begin{tabular}{ll|cccc|cccc}
\hline
& \multirow{2}{*}{\textbf{Models}} & \multicolumn{4}{c}{\textbf{Unseen}} &\multicolumn{4}{c}{\textbf{Seen}} \\
\cline{3-10}
& & Education & Microblog & Science & Subtitles & Laws & News & Spoken & Thesis \\
\hline
1 & \textbf{Vanilla} & 27.52 & 26.05 & 31.58 & 18.32 & 46.69 & 28.67 & 26.44 & 29.00 \\
\hline
2 & \textbf{Vanilla + tag} & 27.36 & 26.11 & 31.53 & 18.25 & 47.13 & 28.75 & 26.71 & 29.19 \\
\hline
3 & \textbf{Meta-MT w/o FT} & 28.76 & 26.41 & 32.41 & 17.38 & 43.74 & 27.31 & 25.98 & 28.11 \\
\hline
4 & \textbf{Meta-Curriculum (LM) w/o FT} & 28.53 & 26.14 & 32.25 & 17.45 & 43.87 & 27.25 & 27.57 & 28.23 \\
\hline
5 & \textbf{RMLNMT w/o FT} & \textbf{30.17} & \textbf{28.42} & \textbf{34.20} & \textbf{19.89} & \textbf{57.54} & \textbf{30.39} & \textbf{28.11} & \textbf{33.20} \\
\hline
\end{tabular}
}
\caption{\label{tab:en2zh_res_domain_robustness}
Domain Robustness: BLEU scores on English $\rightarrow$ Chinese translation tasks.
}
\figlabel{tab:en2zh_res_domain_robustness}
\end{table*}

\begin{table*}
\centering
\resizebox{\textwidth}{!}{
\begin{tabular}{ll|cccc|cccc}
\hline
& \multirow{2}{*}{\textbf{Models}} & \multicolumn{4}{c}{\textbf{Unseen}} &\multicolumn{4}{c}{\textbf{Seen}} \\
\cline{3-10}
& & Education & Microblog & Science & Subtitles & Laws & News & Spoken & Thesis \\
\hline
1 & \textbf{Plain FT} & 27.05 & 26.31 & 32.09 & 17.77 & 47.64 & 28.28 & 25.73 & 28.47 \\
\hline
2 & \textbf{Plain FT + tag} & 27.13 & 26.48 & 32.12 & 17.94 & 47.91 & 28.84 & 26.35 & 29.58 \\
\hline
3 & \textbf{Meta-MT + FT} & 29.33 & 27.48 & 33.12 & 18.77 & 45.21 & 28.43 & 26.82 & 29.20 \\
\hline
4 & \textbf{Meta-Curriculum (LM) + FT} & 28.91 & 27.20 & 33.19 & 18.93 & 45.46 & 28.17 & 27.84 & 29.47 \\
\hline
5 & \textbf{RMLNMT + FT} & \textbf{30.91} & \textbf{28.52} & \textbf{34.51} & \textbf{20.13} & \textbf{57.58} & \textbf{30.42} & \textbf{28.03} & \textbf{32.25} \\
\hline
\end{tabular}
}
\caption{\label{tab:en2zh_res_domain_adaptbility}
Domain Adaptability: BLEU scores on English $\rightarrow$ Chinese translation tasks.
}
\figlabel{tab:en2zh_res_domain_adaptbility}
\end{table*}

\section{Experiments}\label{sec:experiments}
\paragraph{Datasets.}
We experiment with English\textrightarrow German (\textit{en2de}) and English\textrightarrow Chinese (\textit{en2zh}) translation tasks.
For the \textit{en2de} task, we use the same corpora as \citet{Zhan_Liu_Wong_Chao_2021}.
The data consists of corpora in nine domains (Bible, Books, ECB, EMEA, GlobalVoices, JRC, KDE, TED, WMT-News) publicly available on OPUS\footnote{\url{opus.nlpl.eu}} \cite{tiedemann-2012-parallel} and the COVID-19 corpus\footnote{\url{github.com/NLP2CT/Meta-Curriculum}}.
For \textit{en2zh}, we use UM-Corpus \citep{tian-etal-2014-um} containing eight domains: Education, Microblog, Science, Subtitles, Laws, News, Spoken, Thesis.
We use WMT14 (\textit{en2de}) and WMT18 (\textit{en2zh}) corpus published on the WMT website\footnote{\url{http://www.statmt.org}} as our general domain corpora.
We use WMT19 English monolingual corpora to train the LM model so that we can reproduce results from previous work.

\paragraph{Data Preprocessing.}
For English and German, we preprocessed all data with the Moses tokenizer\footnote{\url{github.com/moses-smt/mosesdecoder}} and use sentencepiece\footnote{\url{github.com/google/sentencepiece}} \cite{kudo-richardson-2018-sentencepiece} to encode the corpus with a joint vocabulary, with size 40,000. After that, we filter the sentence longer than 175 tokens and deduplicate the corpus. 
For Chinese, we perform word segmentation using the Stanford Segmenter \citep{tseng-etal-2005-conditional}. 
To have a fair comparison with previous methods \cite{sharaf-etal-2020-meta, Zhan_Liu_Wong_Chao_2021}, we use the same setting, which randomly sub-sampled $\mathcal{D}_{\text {meta-train }}$ and $\mathcal{D}_{\text {meta-test}}$ for each domain with fixed token sizes in order to simulate domain adaptation tasks in low-resource scenarios.
More details for data used in this paper can be found in Appendix~\ref{appendix:datasets}.

\paragraph{Baselines.} We compare \md with the following baselines: 
\begin{itemize}
    \item \textbf{Vanilla}. A standard Transformer-based NMT system trained on the general domains (WMT14 for \textit{en2de}, WMT18 for \textit{en2zh}) and $\mathcal{D}_{\text {meta-train }}$ corpus in seen-domains. We use the $\mathcal{D}_{\text {meta-train }}$ corpus because meta-learning-based methods also use the $\mathcal{D}_{\text {meta-train }}$ corpus, this is a more fair and stronger baseline.
    \item \textbf{Plain fine-tuning}. Fine-tune the vanilla system on support set of $\mathcal{D}_{\text {meta-test }}$ for each individual domain.
    \item \textbf{Tag}. prepend a domain tag to each sentence to indicate what domain it belongs to \citep{kobus-etal-2017-domain}.
    \item \textbf{Meta-MT}. Standard meta-learning approach on domain adaptation task \citep{sharaf-etal-2020-meta}.
    \item \textbf{Meta-Curriculum (LM)}. Meta-learning approach for domain adaptation using LM score as the curriculum to sample the task \citep{Zhan_Liu_Wong_Chao_2021}.
    \item \textbf{Meta-based w/o FT}. This series of experiments uses the meta-learning system prior to adaptation to the specific domain. This can be used to evaluate the domain robustness of meta-based models (see more details in the meta-training part of Section~\ref{sec:meta-learning}).
\end{itemize}

\paragraph{Implementation.}
We use the Transformer model~\citep{vaswani2017attention} as implemented in FairSeq\footnote{\url{github.com/facebookresearch/fairseq}} \citep{ott-etal-2019-fairseq}.
For our word-level domain-mixing modules, we dynamically adjust the network structure according to the number of domains since every domain has its multi-head layers.
Hence, the number of model parameters in the attentive sub-layers of \md is $k$ times the number in the standard transformer ($k$ is the number of seen domains
in the training data).
Following \citet{jiang-etal-2020-multi-domain}, we enlarged the baseline models to have $\sqrt{k}$ times larger embedding dimension, so the baseline has the same number of parameters. This should rule out that the improvements are due to increased parameter count rather than modeling improvements.
For our meta-learning framework, we consider the general meta loss and word-adaptive loss together (as seen in Section~\ref{sec:meta-learning}).
Following \citet{Zhan_Liu_Wong_Chao_2021}, the fine-tuning process in each models is strictly limited to 20 to simulate quick adaptation.
Note that the meta-train stage only uses the seen domain corpus and the unseen domain corpus is only used in the meta-test stage.
More details on hyper-parameters are listed in Appendix~\ref{appendix:model_config}.

\paragraph{Evaluation.}
For a fair comparison with previous work, we use the same data from the support set of $\mathcal{D}_{\text {meta-test}}$ to finetune the model and the same data from the query set of $\mathcal{D}_{\text {meta-test}}$ to evaluate the models.
We measure case-sensitive detokenized BLEU with SacreBLEU\footnote{\url{github.com/mjpost/sacrebleu}} \citep{post-2018-call}; beam search with a beam of size five is used.
Because of the recent criticism of BLEU score \citep{mathur-etal-2020-tangled}, we also evaluate our models using chrF \citep{popovic-2015-chrf} and COMET\footnote{\url{github.com/Unbabel/COMET}} \citep{rei-etal-2020-comet}; the results are listed in Appendix~\ref{appendix:evaluations}.

\paragraph{Domain Robustness.}
Domain robustness shows the effectiveness of the model both in seen and unseen domains. Hence, we use the model without fine-tuning to evaluate the domain robustness.

\paragraph{Domain Adaptability.}
We evaluate the domain adaptability by testing that the model quickly adapts to new domains using just hundreds of in-domain parallel sentences.
Therefore, we fine-tune the models on a small amount of domain-specific data.

\paragraph{Cross-Domain Robustness.}
To better show the cross-domain robustness of \md, we use the fine-tuned model of one specific domain to generate the translation for other domains.
More formally, given $k$ domains, we use the fine-tuned model $M_{J}$ with the domain label of $J$ to generate the translation of $k$ domains.

\begin{table}[ht]
    \centering
    \begin{tabular}{l|c}
        \hline
        \textbf{Methods} & \textbf{Avg} \\
        \hline
        Meta-MT & -1.97 \\
        Meta-Curriculum (LM) &  -0.96 \\
        Meta-Curriculum (cls) & -0.98 \\
        \md & \textbf{2.64} \\
        \hline
    \end{tabular}
    \caption{The average improvement over vanilla baseline.}
    \label{tab:robustness}
    \tablabel{tab:robustness}
\end{table}

\begin{table*}[t]
\centering
\resizebox{\textwidth}{!}{
\begin{tabular}{l|ccccc|ccccc}
\hline
\multirow{2}{*}{\textbf{Classifier}} & \multicolumn{5}{c}{\textbf{Unseen}} &\multicolumn{5}{c}{\textbf{Seen}} \\
\cline{2-11}
& Covid & Bible & Books & ECB & TED & EMEA & Globalvoices & JRC & KDE & WMT \\
\hline
\textbf{CNN} & 24.12 & 13.57 & 12.74 & 30.31 & 28.14 & 46.12 & 25.17 & 50.52 & 31.15 & 26.34 \\
\hline
\textbf{BERT-many-labels} & 25.89 & 14.77 & 13.71 & \textbf{32.10} & 29.28 & \textbf{47.41} & \textbf{26.70} & 51.34 & 32.76 & 28.17 \\
\hline
\textbf{BERT-2-labels} & 26.10 & 14.85 & 13.58 & 31.99 & 29.17 & 46.80 & 26.46 & \textbf{51.56} & 32.83 & 28.37 \\
\hline
\textbf{mBERT-many-labels} & 26.10 & 14.73 & 13.69 & 31.93 & 29.11 & 47.02 & 26.33 & 51.13 & 32.69 & 27.91 \\
\hline
\textbf{mBERT-2-labels} & \textbf{26.53 }& \textbf{15.37} & \textbf{13.71} & 31.97 & \textbf{29.47} & 47.02 & 26.55 & 51.13 & \textbf{32.88} & \textbf{28.37} \\
\hline
\end{tabular}
}
\caption{\label{tab:diff_cls}Different classifier: BLEU scores on the English $\rightarrow$ German translation task.
}
\figlabel{tab:diff_cls}
\end{table*}

\begin{table*}[t]
\centering
\resizebox{\textwidth}{!}{
\begin{tabular}{l|ccccc|ccccc}
\hline
\multirow{2}{*}{\textbf{Sampling Strategy}} & \multicolumn{5}{c}{\textbf{Unseen}} &\multicolumn{5}{c}{\textbf{Seen}} \\
\cline{2-11}
& Covid & Bible & Books & ECB & TED & EMEA & Globalvoices & JRC & KDE & WMT \\
\hline
\textbf{Token-based sampling} & 25.30 & 11.38 & 12.70 & 31.61 & 28.01 & 47.51 & 26.50 & \textbf{51.31} & 32.88 & 28.03 \\
\hline
\textbf{Balance sampling} & \textbf{25.47} & \textbf{11.51} & \textbf{12.79} & \textbf{32.08} & \textbf{28.98} & \textbf{47.64} & \textbf{26.58} & 51.25 & \textbf{32.91} & \textbf{28.07} \\
\hline
\end{tabular}
}
\caption{\label{tab:diff_samp}Different sampling strategy: BLEU scores on the English $\rightarrow$ German translation task.
}
\figlabel{tab:diff_samp}
\end{table*}

\begin{table*}[!h]
\centering
\resizebox{\textwidth}{!}{
\begin{tabular}{l|ccccc|ccccc}
\hline
\multirow{2}{*}{\textbf{Finetune Strategy}} & \multicolumn{5}{c}{\textbf{Unseen}} &\multicolumn{5}{c}{\textbf{Seen}} \\
\cline{2-11}
& Covid & Bible & Books & ECB & TED & EMEA & Globalvoices & JRC & KDE & WMT \\
\hline
\textbf{FT-unseen} & 25.23 & 13.18 & 12.73 & \textbf{32.45} & 28.41 & 46.35 & 25.83 & 50.85 & 32.30 & 26.88 \\
\hline
\textbf{FT-seen} & 24.58 & 11.73 & 12.57 & 30.79 & 27.29 & 46.58 & 25.73 & 50.91 & 31.78 & 26.51 \\
\hline
\textbf{FT-all} & 15.00 & 7.77 & 9.06 & 21.33 & 16.98 & 24.69 & 14.63 & 27.59 & 12.77 & 15.75 \\
\hline
\textbf{FT-specific} & \textbf{26.53} & \textbf{15.37} & \textbf{13.71} & 31.97 & \textbf{29.47} & \textbf{47.02} & \textbf{26.33} & \textbf{51.13} & \textbf{32.83} & \textbf{28.37} \\
\hline
\end{tabular}
}
\caption{\label{tab:diff_finetune}Different fine-tuning strategy: BLEU scores on the English $\rightarrow$ German translation task.
}
\figlabel{tab:diff_finetune}
\end{table*}

\section{Results}\label{sec:results}

Table~\ref{tab:en2de_res_domain_robustness} and Table~\ref{tab:en2zh_res_domain_robustness} show the domain robustness for English$\rightarrow$German and English$\rightarrow$Chinese respectively. Table~\ref{tab:en2de_res_domain_adapt} and Table~\ref{tab:en2zh_res_domain_adaptbility} show the domain adaptability on both translation task.

\paragraph{Domain Robustness.}
As seen in Table~\ref{tab:en2de_res_domain_robustness} and Table~\ref{tab:en2zh_res_domain_robustness}, \md shows the best domain robustness compared with other models both in seen and unseen domains.
In addition, the traditional meta-learning approach (Meta-MT, Meta-Curriculum) without fine-tuning is even worse than the standard transformer model in seen domains. This phenomenon is our motivation for improving the robustness of traditional meta-learning based approach.
In other words, we cannot be sure whether the improvement of the meta-based method is due to the domain adaptability of meta-learning or the robustness of the teacher model.
Note this setup differs from the previous work \citep{sharaf-etal-2020-meta, Zhan_Liu_Wong_Chao_2021} because we included the $\mathcal{D}_{\text {meta-train }}$ data to the vanilla system to insure all systems in the table use the same training data.\footnote{We also confirmed with \citet{Zhan_Liu_Wong_Chao_2021} via email that they did not deduplicate the corpus, which is another reason for the difference between our results and their results.}
Interestingly, the translation quality in the \emph{WMT} domain is also improved which is different than \cite{Zhan_Liu_Wong_Chao_2021}. They explain that their methods achieve maximum robustness on the \emph{WMT} domain, while our results demonstrate that our model can further improve robustness even when trained on
the same domain as the pre-trained model.

\paragraph{Domain Adaptability.}
From Tables~\ref{tab:en2de_res_domain_adapt} and~\ref{tab:en2zh_res_domain_adaptbility}, we observe that the traditional meta-learning approach shows high adaptability to unseen domains but fails on seen domains due to limited domain robustness.
In contrast, \md shows its domain adaptability both in seen and unseen domains, and maintains the domain robustness simultaneously.
Compared with \md, the traditional meta-learning approach show more improvement between the \emph{w/o FT} model and \emph{FT} model. For example, \emph{Meta-MT} and \emph{Meta-Curriculum (LM)} obtains $1.32$ and $2.19$ BLEU score improvement after finetuning on the \emph{ECB} domain; improvement from \emph{\md} only got $0.55$. This phenomenon meets our expectations since \emph{\md}  without finetuning is already strong enough due to the domain robustness of word-level domain mixing.
In other words, the improvement of the traditional meta-learning approach is to some extent due to the unrobustness of the model.

\paragraph{Cross-Domain Robustness.}
Table~\ref{tab:robustness} reports the average difference of $k \times k$ BLEU scores; a larger positive value means a more robust model. 
We observed that the plain meta-learning based methods have a negative value, which means the performance gains in the specific domains come at the cost of performance decreases in other domains. In other words, the model is not domain robust enough.
In contrast, \md has a positive difference with the vanilla system, showing that the model is robust.
The specific BLEU scores are shown in Figure~\ref{Fig:robust_res} of Appendix~\ref{appendix:cross-domain-robustness}.

The results of both domain robustness and domain adaptability are consistent for the chrF and COMET evaluation metrics (see more details in Tables~\ref{tab:en2de_res_chrf} and~\ref{tab:en2de_res_comet} of Appendix~\ref{appendix:evaluations}).

\section{Analysis}

In this section, we conduct additional experiments to better understand the strengths of \md.
We analyze the contribution of different components in \md, through an ablation study.

\paragraph{Different classifiers.}
We evaluate the impact of different classifiers on translation performance. The main results are as shown in Table~\ref{tab:diff_cls} (see more details in Appendix~\ref{appendix:df_cls}).
We observed that the performance of \md is not directly proportional to the accuracy of the classifier.
In other words, slightly higher classification accuracy does not lead to better BLEU scores.
This is because the accuracy of the classifier is close between BERT-based models and the primary role of the classifier is to construct the curriculum for splitting the tasks.
When we use a significantly worse classifier, i.e., the CNN in our experiments, the overall performance of \md is worse than the BERT-based classifier.

\paragraph{Balanced sampling vs. Token-based sampling.}
Plain meta-learning uses a token-based sampling strategy to split sentences into small tasks.
However, the token-based strategy could cause unbalanced domain distribution in some tasks, especially in the early stage of training due to domain mismatches (see the discussion of balanced sampling in Section~\ref{sec:meta-learning}).
To address this issue, we proposed to balance the domain distribution after splitting the task.
Table~\ref{tab:diff_samp} shows that our methods can result in small improvements in performance.
For example, in the \textit{TED} domain, BLEU was 28.01 with token-based sampling, but with the balanced sampling strategy BLEU was 28.98.
We keep the same number of tasks to have a fair comparison with previous methods.

\paragraph{Different fine-tuning strategies.}
As described in Section~\ref{sec:word-domain-mixing}, the model for each domain has its own multi-head and feed-forward layers.
During the fine-tuning stage of \md, we devise four strategies:
i) \textbf{FT-unseen}: fine-tuning using all unseen domain corpora;
ii) \textbf{FT-seen}: fine-tuning using all seen domain corpora;
iii) \textbf{FT-all}: fine-tuning using all out-of-domain corpora (seen and unseen domains);
iv) \textbf{FT-specific}: using the specific domain corpus to fine-tune the specific models.
The results are shown in Table~\ref{tab:diff_finetune}.
\textit{FT-specific} obtains robust results among all the strategies.
Although other strategies outperform \textit{FT-specific} in some domains, \textit{FT-specific} is robust across all domains.
Furthermore, \textit{FT-specific} is the fairest comparison because it uses only a specific domain corpus to fine-tune, which is the same as the baseline systems.

\section{Related Work}

\paragraph{Domain Adaptation for NMT.}
Current approaches can be categorized into two groups by granularity:
From a sentence-level perspective, researchers either use data selection methods \citep{moore-lewis-2010-intelligent, axelrod-etal-2011-domain} to select the training data that is similar to out-of-domain parallel corpora or train a classifier \citep{riess21:mtsummit} or utilize a language model \citep{wang-etal-2017-instance, Zhan_Liu_Wong_Chao_2021} to better weight the sentences.
From a word-level perspective, researchers try to model domain distribution at the word level, since a word in a sentence can be related to more domains than just the sentence domain
\citep{zeng-etal-2018-multi,yan-etal-2018-word,hu-etal-2019-domain-adaptation, sato-etal-2020-vocabulary, jiang-etal-2020-multi-domain}.

\paragraph{Curriculum Learning for NMT.}
Curriculum learning \citep{bengio2009curriculum} starts with easier tasks and then progressively gain experience to process more complex tasks, which has proved to be useful in NMT domain adaptation.
\citet{stojanovski-fraser-2019-improving} utilize curriculum learning to improve anaphora resolution in NMT systems.
\citet{zhang-etal-2019-curriculum} and \citet{Zhan_Liu_Wong_Chao_2021} use a language model to compute a similarity score between domains, from which a curriculum is devised for adapting NMT systems to specific domains from general domains.

\paragraph{Meta-Learning for NMT.}
\citet{gu-etal-2018-meta} apply model-agnostic meta-learning (MAML; \citealp{finn2017model}) to NMT. They show that MAML effectively improves low-resource NMT.
\citet{li2020metamt}, \citet{sharaf-etal-2020-meta} and \citet{Zhan_Liu_Wong_Chao_2021} propose to formulate the problem of low-resource domain adaptation in NMT as a meta-learning problem: the model learns to quickly adapt to an unseen new domain from a general domain.

\section{Conclusion}
We presented \texttt{RMLNMT}, a robust meta-learning framework for low-resource NMT domain adaptation reaching both high domain adaptability and domain robustness (both in the seen domains and unseen domains).
We found that domain robustness dominates the results compared to domain adaptability in meta-learning based approaches.
The results show that \md works best in setups that require high robustness in low-resource scenarios. 

\section*{Acknowledgement}
We thank Mengjie Zhao for the helpful comments.
This work was supported by funding to Wen Lai's PhD research from LMU-CSC (China Scholarship Council) Scholarship Program. This work has received funding from the European Research Council under the European Union’s Horizon $2020$ research and innovation program (grant agreement 
\#$640550$). This work was also supported by the DFG (grant FR $2829$/$4$-$1$). 
The work at CUNI was supported by the European Commission via its Horizon 2020 research and innovation programme (870930).

\bibliography{custom}
\bibliographystyle{acl_natbib}

\appendix

\section{Appendix}

\subsection{Datasets}
\label{appendix:datasets}
For the OPUS corpus used in the English $\rightarrow$ German translation task, we deduplicated the corpus, which is different from \cite{Zhan_Liu_Wong_Chao_2021} and is the main reason that we cannot reproduce the results in the original paper. 
The statistics of the original OPUS are shown in Table~\ref{tab:data_filter}.
The seen domains (EMEA, Globalvoices, JRC, KDE, WMT) contain a lot of duplicated sentences.
The scores in the original paper are too high because the $\mathcal{D}_{\text {meta-train}}$ dataset overlaps with some sentences in $\mathcal{D}_{\text {meta-test}}$.

\begin{table}[h]
    \centering
    \begin{tabular}{c|r|r}
        \hline
        \textbf{Corpus} & \textbf{Original} & \textbf{Deduplicated} \\
        \hline
        Covid & 3,325 & 3,312 \\
        \hline
        Bible & 62,195 & 61,585 \\
         \hline
        Books & 51,467 & 51,106 \\
         \hline
        ECB & 113,116 & 113,081 \\
         \hline
        TED & 143,830 & 142,756 \\
         \hline
        EMEA & 1,103,807 & 360,833 \\
         \hline
        Globalvoices & 71,493 & 70,519 \\
         \hline
        JRC & 717,988 & 503,789 \\
         \hline
        KDE & 223,672 & 187,918 \\
         \hline
        WMT & 45,913 & 34,727 \\
        \hline
    \end{tabular}
    \caption{Data statistic (sentences) of the original corpus for English$\rightarrow$German translation task}
    \label{tab:data_filter}
\end{table}

\begin{table}[t]
    \centering
    \scalebox{0.8}{
    \begin{tabular}{cccccc}
        \hline
        & \multicolumn{2}{c}{$\mathcal{D}_{\text {meta-train}}$} & & \multicolumn{2}{c}{$\mathcal{D}_{\text {meta-test}}$} \\
        \cline{2-3} \cline{5-6}
        & \textbf{Support} & \textbf{Query} && \textbf{Support} & \textbf{Query} \\
        \hline
        \textbf{Covid} & / & / && 309 & 612 \\
        \hline
        \textbf{Bible} & / & / && 280 & 548 \\
         \hline
        \textbf{Books} & / & / && 304 & 637 \\
         \hline
        \textbf{ECB} & / & / && 295 & 573 \\
         \hline
        \textbf{TED} & / & / && 390 & 772 \\
         \hline
        \textbf{EMEA} & 14856 & 29668 && 456 & 975\\
         \hline
        \textbf{Globalvoices} & 11686 & 23319 && 368 & 699 \\
         \hline
        \textbf{JRC} & 7863 & 15769 && 254 & 519 \\
         \hline
        \textbf{KDE} & 24078 & 48284 && 756 & 1510 \\
         \hline
        \textbf{WMT} & 10939 & 21874 && 334 & 704\\
        \hline
    \end{tabular}
    }
    \caption{Data statistic (sentences) of the meta-learning stage for English$\rightarrow$German translation task}
    \label{tab:data_en2de_meta}
\end{table}

\begin{table}[h]
    \centering
    \scalebox{0.8}{
    \begin{tabular}{cccccc}
        \hline
        & \multicolumn{2}{c}{$\mathcal{D}_{\text {meta-train}}$} & & \multicolumn{2}{c}{$\mathcal{D}_{\text {meta-test}}$} \\
        \cline{2-3} \cline{5-6}
        & \textbf{Support} & \textbf{Query} && \textbf{Support} & \textbf{Query} \\
        \hline
        \textbf{Education} & / & / && 395 & 785 \\
        \hline
        \textbf{Microblog} & / & / && 358 & 721 \\
         \hline
        \textbf{Science} & / & / && 392 & 852 \\
         \hline
        \textbf{Subtitles} & / & / && 612 & 1219 \\
         \hline
        \textbf{Laws} & 6379 & 13001 && 197 & 416 \\
         \hline
        \textbf{News} & 9004 & 18362 && 281 & 536 \\
         \hline
        \textbf{Spoken} & 18270 & 36569 && 571 & 1148 \\
         \hline
        \textbf{Thesis} & 8914 & 17883 && 298 & 547 \\
         \hline
    \end{tabular}
    }
    \caption{Data statistic (sentences) of the meta-learning stage for English$\rightarrow$Chinese translation task}
    \label{tab:data_en2zh_meta}
\end{table}

For the meta-learning phase, to have a fair comparison with previous methods, we use the same setting. We random split 160 tasks and 10 tasks respectively in $\mathcal{D}_{\text {meta-train}}$ and $\mathcal{D}_{\text {meta-test}}$ to simulate the low-resource scenarios. For each task, the token amount of support set and query set is a strict limit to $8K$ and $16K$. $\mathcal{D}_{\text {meta-dev}}$ corpus is limited to 5000 sentences for each domain. Table~\ref{tab:data_en2de_meta} and Table~\ref{tab:data_en2zh_meta} shows the detailed statistics of the English $\rightarrow$ German and English $\rightarrow$ Chinese tasks.

\subsection{Model Configuration}
\label{appendix:model_config}
We use the \texttt{Transformer Base} architecture \cite{vaswani2017attention} as implemented in fairseq \cite{ott-etal-2019-fairseq}. We use the standard Transformer architecture with dimension 512, feed-forward layer 2048, 8 attention heads, 6 encoder layers and 6 decoder layers. For optimization, we use the Adam optimizer with a learning rate of $5\cdot 10^{-5}$. To prevent overfitting, we applied a dropout of 0.3 on all layers. 
The number of warm-up steps was set to 4000.
At the time of inference, a beam search of size 5 is used to balance the decoding time and accuracy of the search. 

For the word-level domain-mixing model, we use the same setting as \citet{jiang-etal-2020-multi-domain}. The number of parameters of our model is dynamically adjusted with the domain numbers and $k$ times higher than standard model architecture, since every domain has its multi-head attention layer and feed-forward layer.
To have a fair comparison between baselines, we enlarged the baseline models to have $\sqrt{k}$ times larger embedding dimension, so the baseline has the same number of parameters.

\begin{table}[t]
    \centering
    \begin{tabular}{l|c}
        \hline
        \textbf{Classifier} & \textbf{Acc(\%)} \\
        \hline
        CNN & 74.91\% \\
        BERT: many-labels &  96.12\% \\
        BERT: 2-labels & 95.35\% \\
        mBERT: many-labels & 95.41\% \\
        mBERT: 2-labels & 95.26\% \\
        \hline
    \end{tabular}
    \caption{The accuracy of the different classifiers.}\label{tab:acc_cls}
    \tablabel{tab:acc_cls}
\end{table}

\subsection{Different classifiers}
\label{appendix:df_cls}
With a general in-domain corpus and some out-of-domain corpora, we train five classifiers.
We experiment with two different labeling schemes: \texttt{2-labels} where we distinguish only two classes: \textit{out-of-domain} and \textit{in-domain};
\texttt{many-labels} where sentences are labeled with the respective domain labels.
Further, we experiment with two variants of the BERT model:
first, we use monolingual English BERT on the source side only, and second, we use multilingual BERT (mBERT) to classify the parallel sentence pairs.
For further comparison, we include also a CNN-based classifier \citep{kim-2014-convolutional}.
We present the accuracy of the English-German domain classifier in \tabref{tab:acc_cls}.

\subsection{Cross-Domain Robustness}
\label{appendix:cross-domain-robustness}
\begin{figure*}[t]
\includegraphics[width=0.99\textwidth]{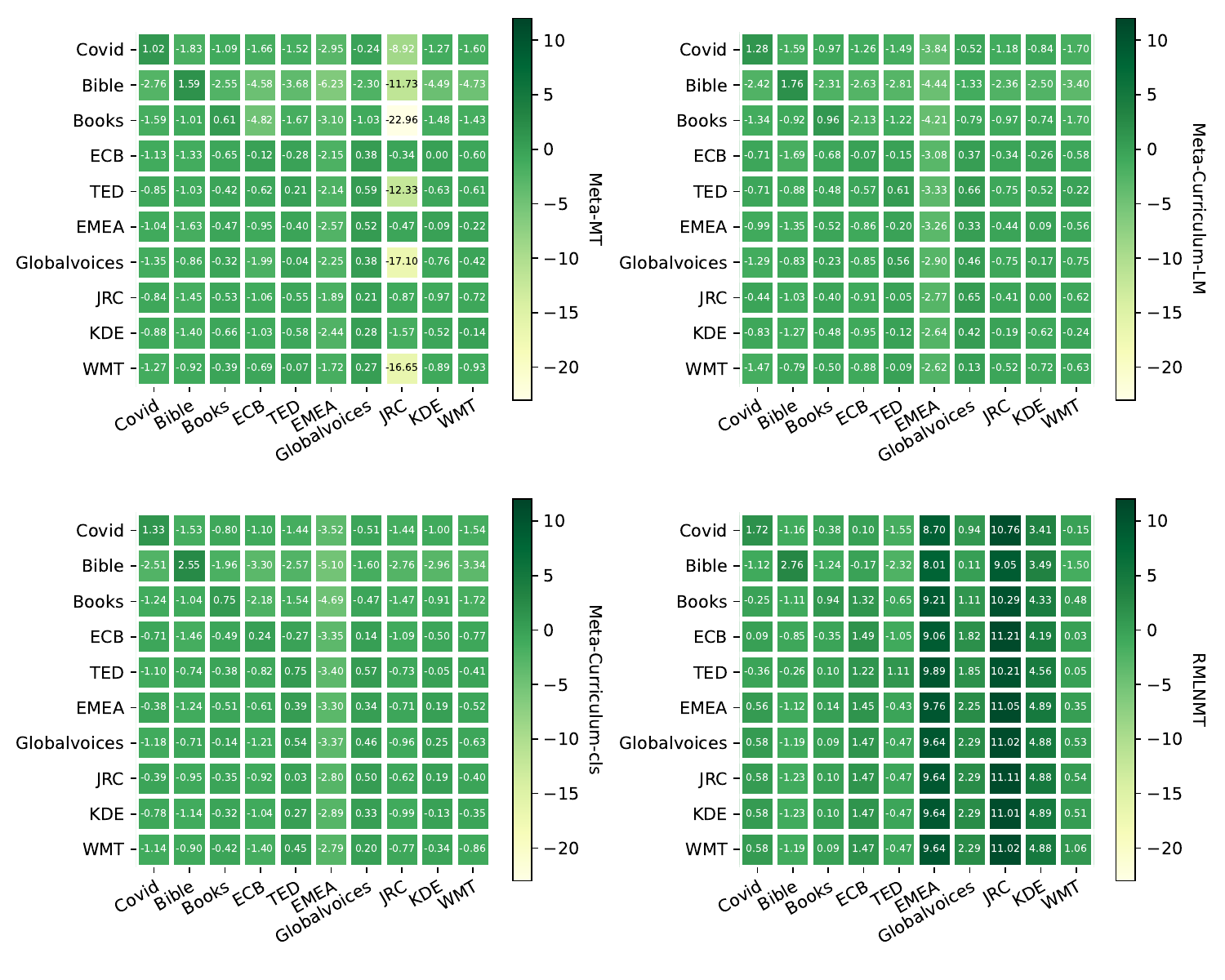}
\caption{BLEU scores for one specific finetuned model on other domains for en2de translation.}\label{Fig:robust_res}
\end{figure*}

In Figure~\ref{Fig:robust_res} we show the detailed results ($k \times k$ scores) of cross-domain robustness.

\subsection{Evaluations}\label{appendix:evaluations}

In addition to BLEU, we also use chrF \citep{popovic-2015-chrf} and COMET \citep{rei-etal-2020-comet} as evaluation metrics. Table~\ref{tab:en2de_res_chrf} and Table~\ref{tab:en2de_res_comet} show the results. Consistently with the BLEU score (Tables~\ref{tab:en2de_res_domain_robustness} and Table~\ref{tab:en2de_res_domain_adapt}), we observed that \md is more effective than all previous methods.

\begin{table*}[htbp]
\centering
\scalebox{0.75}{
\begin{tabular}{ll|ccccc|ccccc}
\hline
& \multirow{2}{*}{\textbf{Models}} & \multicolumn{5}{c}{\textbf{Unseen}} &\multicolumn{5}{c}{\textbf{Seen}} \\
\cline{3-12}
& & Covid & Bible & Books & ECB & TED & EMEA & Globalvoices & JRC & KDE & WMT \\
\hline
\multirow{2}{*}{1} & \textbf{Vanilla} & 0.550 & 0.418 & 0.385 & 0.538 & 0.542 & 0.599 & 0.536 & 0.614 & 0.525 & 0.558 \\
& \textbf{Plain FT} & 0.555 & 0.423 & 0.388 & 0.540 & 0.548 & 0.600 & 0.536 & 0.618 & 0.528 & 0.558 \\
\hline
\multirow{2}{*}{2} & \textbf{Vanilla + tag} & 0.555 & 0.418 & 0.384 & 0.540 & 0.544 & 0.657 & 0.545 & 0.627 & 0.531 & 0.558 \\
& \textbf{Plain FT + tag} & 0.562 & 0.423 & 0.388 & 0.540 & 0.549 & 0.602 & 0.536 & 0.694 & 0.547 & 0.561 \\
\hline
\multirow{2}{*}{3} & \textbf{Meta-MT w/o FT} & 0.545 & 0.410 & 0.382 & 0.498 & 0.538 & 0.532 & 0.531 & 0.610 & 0.464 & 0.553 \\
& \textbf{Meta-MT + FT} & 0.566 & 0.432 & 0.390 & 0.542 & 0.556 & 0.582 & 0.538 & 0.613 & 0.522 & 0.552 \\
\hline
\multirow{2}{*}{4} &\textbf{Meta-Curriculum (LM) w/o FT} & 0.548 & 0.412 & 0.384 & 0.523 & 0.543 & 0.560 & 0.536 & 0.611 & 0.521 & 0.554 \\
& \textbf{Meta-Curriculum (LM) + FT} & \textbf{0.567} & 0.434 & 0.395 & 0.544 & 0.548 & 0.572 & 0.539 & 0.615 & 0.522 & 0.553 \\
\hline
\multirow{2}{*}{5}& \textbf{RMLNMT w/o FT} & 0.555 & 0.405 & 0.388 & 0.557 & 0.544 & 0.656 & 0.552 & 0.702 & 0.574 & 0.561 \\
& \textbf{RMLNMT + FT} & 0.562 & \textbf{0.451} & \textbf{0.395} & \textbf{0.558} & \textbf{0.560} & \textbf{0.656} & \textbf{0.552} & \textbf{0.702} & \textbf{0.574} & \textbf{0.561} \\
\hline
\end{tabular}
}
\caption{\label{tab:en2de_res_chrf}chrF scores on the English $\rightarrow$ German translation task.
}
\figlabel{tab:en2de_chrF_res}
\end{table*}

\begin{table*}[htbp]
\centering
\scalebox{0.70}{
\begin{tabular}{ll|ccccc|ccccc}
\hline
& \multirow{2}{*}{\textbf{Models}} & \multicolumn{5}{c}{\textbf{Unseen}} &\multicolumn{5}{c}{\textbf{Seen}} \\
\cline{3-12}
& & Covid & Bible & Books & ECB & TED & EMEA & Globalvoices & JRC & KDE & WMT \\
\hline
\multirow{2}{*}{1} & \textbf{Vanilla} & 0.4967 & -0.1250 & -0.2225 & 0.3276 & 0.3400 & 0.3096 & 0.3199 & 0.5430 & 0.1836 & 0.4326 \\
& \textbf{Plain FT} & 0.5066 & -0.1105 & -0.1985 & 0.3315 & 0.3553 & 0.3177 & 0.3276 & 0.5492 & 0.1813 & 0.4392 \\
\hline
\multirow{2}{*}{2} & \textbf{Vanilla + tag} & 0.4970 & -0.1250 & -0.2228 & 0.3277 & 0.3401 & 0.3176 & 0.3291 & 0.5495 & 0.1846 & 0.4311 \\
& \textbf{Plain FT + tag} & 0.5078 & -0.1105 & -0.1981 & 0.3315 & 0.3553 & 0.3179 & 0.3341 & 0.5572 & 0.1973 & 0.4398 \\
\hline
\multirow{2}{*}{3} & \textbf{Meta-MT w/o FT} & 0.4850 & -0.1454 & -0.2228 & 0.0953 & 0.3506 & 0.0524 & 0.2985 & 0.5319 & 0.1304 & 0.4137 \\
& \textbf{Meta-MT + FT} & 0.5175 & -0.0650 & -0.1878 & 0.3466 & 0.3824 & 0.2678 & 0.3189 & 0.5509 & 0.1316 & 0.4161 \\
\hline
\multirow{2}{*}{4} &\textbf{Meta-Curriculum (LM) w/o FT} & 0.4879 & -0.1365 & -0.2122 & 0.2568 & 0.3751 & 0.1968 & 0.3273 & 0.5246 & 0.0962 & 0.4206 \\
& \textbf{Meta-Curriculum (LM) + FT} & \textbf{0.5347} & -0.0604 & -0.1773 & 0.3460 & 0.3729 & 0.2366 & 0.3141 & 0.5430 & 0.1467 & 0.4128 \\
\hline
\multirow{2}{*}{5}& \textbf{RMLNMT w/o FT} & 0.4943 & -0.1956 & -0.2179 & 0.3580 & 0.3394 & 0.4026 & 0.3769 & 0.6797 & 0.3014 & 0.4255 \\
& \textbf{RMLNMT + FT} & 0.5302 & \textbf{-0.0543} & \textbf{-0.1610} & \textbf{0.3547} & \textbf{0.3867} & \textbf{0.4046} & \textbf{0.3771} & \textbf{0.6797} & \textbf{0.3015} & \textbf{0.4256} \\
\hline
\end{tabular}
}
\caption{\label{tab:en2de_res_comet}COMET scores on the English $\rightarrow$ German translation task.
}
\figlabel{tab:en2de_comet_res}
\end{table*}

\end{document}